\documentclass{article}

\PassOptionsToPackage{numbers, compress}{natbib}

\usepackage[preprint]{neurips_2025}

\usepackage{enumitem}
\usepackage[utf8]{inputenc} % allow utf-8 input
\usepackage[T1]{fontenc}    % use 8-bit T1 fonts
\usepackage{hyperref}       % hyperlinks
\usepackage{url}            % simple URL typesetting
\usepackage{booktabs}       % professional-quality tables
\usepackage{amsfonts}       % blackboard math symbols
\newcommand{\R}{\mathbb{R}}
\usepackage{nicefrac}       % compact symbols for 1/2, etc.
\usepackage{microtype}      % microtypography
\usepackage{xcolor}         % colors
\usepackage{amsmath}
\usepackage{amssymb}
\usepackage{cleveref} 
\usepackage{amsfonts} % For mathbb
\usepackage{graphicx}
\expandafter\def\expandafter\normalsize\expandafter{%
    \normalsize%
    \setlength\abovedisplayskip{4pt}%
    \setlength\belowdisplayskip{4pt}%
    \setlength\abovedisplayshortskip{2pt}%
    \setlength\belowdisplayshortskip{2pt}%
}
\newcommand{\E}{\mathbb{E}} % Expectation
\newcommand{\Ls}{\mathcal{L}} % Loss symbol
\newcommand{\vect}[1]{\mathbf{#1}} % Vectors (bold)
\newcommand{\params}[1]{\theta_{#1}} % Parameters theta_subscript
\DeclareMathOperator*{\MAE}{MAE}
\DeclareMathOperator*{\Lone}{L1} % L1 norm/loss
\DeclareMathOperator*{\Huber}{Huber}
\DeclareMathOperator*{\KL}{KL} % KL Divergence
\DeclareMathOperator*{\softplus}{softplus} % Softplus function
\DeclareMathOperator*{\BetaDist}{Beta} % Beta distribution
\DeclareMathOperator*{\lgamma}{lgamma} % Log Gamma function
\title{EvidenceMoE: A Physics-Guided Mixture-of-Experts with Evidential Critics for Advancing Fluorescence Light Detection and Ranging in Scattering Media}

\author{
  Ismail Erbas \\
  Center for Modeling, Simulation, \& Imaging in Medicine\\
  Rensselaer Polytechnic Institute\\
  Troy, NY \\
  \texttt{erbasi@rpi.edu} \\
    \And
Ferhat Demirkiran \\
Department of Computer Science\\
  University at Albany \\
  Albany, NY  \\
  \texttt{fdemirkiran@albany.edu} \\
   \And
  Karthik Swaminathan \\
  IBM T.J. Watson Research Center \\
  Yorktown Heights, NY  \\
  \texttt{kvswamin@us.ibm.com} \\
  \And
  Naigang Wang \\
  IBM T. J. Watson Research Center \\
  Yorktown Heights, NY \\
  \texttt{nwang@us.ibm.com} \\
   \And
  Navid I. Nizam \\
  Center for Modeling, Simulation, \& Imaging in Medicine\\
  Rensselaer Polytechnic Institute \\
   Troy, NY \\
  \texttt{nizamn@rpi.edu} \\
    \And
  Stefan T. Radev \\
  Center for Modeling, Simulation, \& Imaging in Medicine\\
  Rensselaer Polytechnic Institute \\
  Troy, NY \\
  \texttt{radevs@rpi.edu} \\
   \And
  Kaoutar El Maghraoui \\
  IBM T.J. Watson Research Center \\
  Yorktown Heights, NY  \\
  \texttt{kelmaghr@us.ibm.com} \\
   \And
  Xavier Intes \\
  Center for Modeling, Simulation, \& Imaging in Medicine\\
  Rensselaer Polytechnic Institute \\
   Troy, NY \\
  \texttt{intesx@rpi.edu} \\
 \And
  Vikas Pandey \\
  Center for Modeling, Simulation, \& Imaging in Medicine\\
  Rensselaer Polytechnic Institute \\
   Troy, NY \\
  \texttt{pandev2@rpi.edu} \\
}
\DeclareUnicodeCharacter{0301}{\'{}}

\begin{document}
\maketitle

\begin{abstract}

Fluorescence LiDAR (FLiDAR), a Light Detection and Ranging (LiDAR) technology employed for distance and depth estimation across medical, automotive, and other fields, encounters significant computational challenges in scattering media. The complex nature of the acquired FLiDAR signal, particularly in such environments, makes isolating photon time-of-flight (related to target depth) and intrinsic fluorescence lifetime exceptionally difficult, thus limiting the effectiveness of current analytical and computational methodologies. To overcome this limitation, we present a Physics-Guided Mixture-of-Experts (MoE) framework tailored for specialized modeling of diverse temporal components. In contrast to the conventional MoE approaches our expert models are informed by underlying physics, such as the radiative transport equation governing photon propagation in scattering media. Central to our approach is EvidenceMoE, which integrates Evidence-Based Dirichlet Critics (EDCs). These critic models assess the reliability of each expert's output by providing per-expert quality scores and corrective feedback. A Decider Network then leverages this information to fuse expert predictions into a robust final estimate adaptively. We validate our method using realistically simulated Fluorescence LiDAR (FLiDAR) data for non-invasive cancer cell depth detection generated from photon transport models in tissue. Our framework demonstrates strong performance, achieving a normalized root mean squared error (NRMSE) of 0.030 for depth estimation and 0.074 for fluorescence lifetime.

\end{abstract}

\section{Introduction}
\begin{figure}[h!]
\centering
\includegraphics[width=0.6\textwidth]{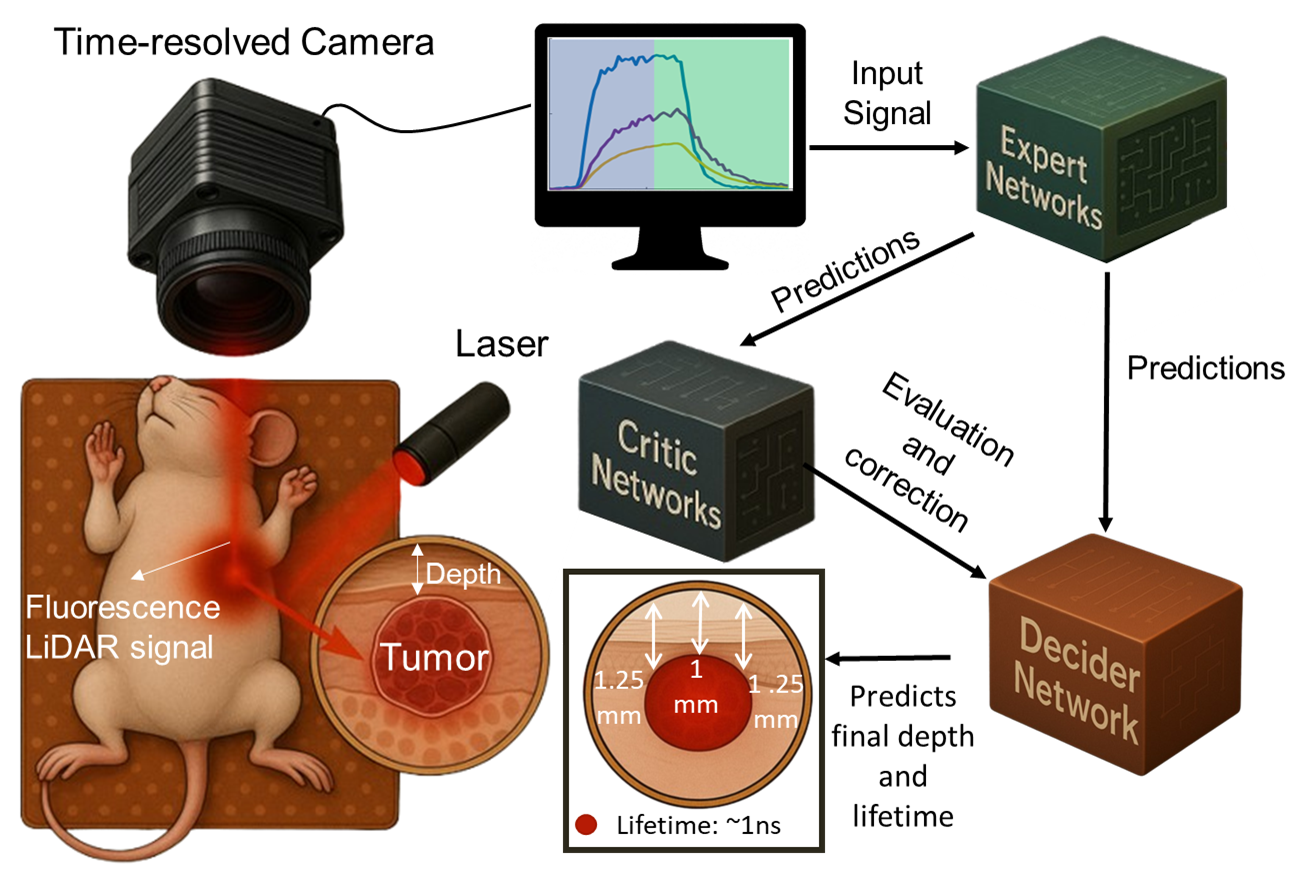}
\caption{\textbf{End-to-end EvidenceMoE Workflow for FLiDAR-based Tumor Lifetime and Depth Estimation.} FLiDAR signal acquisition begins with laser interaction with a fluorescent tumor target. The resulting emitted/scattered photons are captured by a time-resolved camera, yielding a complex temporal signal. EvidenceMoE architecture processes the captured signal to ultimately yield estimations for tumor depth and fluorescence lifetime.}
\label{fig:overall}
\vspace{-0.2in}
\end{figure}

Light Detection and Ranging (LiDAR) stands as a pivotal technology across a wide array of applications, from autonomous navigation to environmental mapping \citep{ma2024review,karim2024application}. LiDAR enables precise depth estimation by measuring the time-of-flight (TOF) of emitted and subsequently detected photons. %The utility of LiDAR spans a wide array of applications, from autonomous navigation to environmental mapping \citep{ma2024review,karim2024application}. 
However, traditional LiDAR becomes less accurate in scattering media, such as biological tissues, fog, or turbid water, where photons deviate from their original paths due to interactions with the media \citep{ma2024review}. 
%LiDAR presents a significant opportunity in fluorescence imaging, enabling tasks such as fluorophore depth estimation within scattering media and the simultaneous assessment of their micro-environmental parameters via Fluorescence Lifetime Imaging (FLI), an intrinsic property of these molecules \citep{smith2020macroscopic}. 
In contrast, Fluorescence LiDAR (FLiDAR) enables effective depth estimation of fluorophores (fluorescent particles) within these scattering media. Further, the intrinsic lifetime properties of these fluorophores can also be used to distinguish micro-environmental differences within the media (such as tissue boundaries), which is termed as Fluorescence Lifetime Imaging (FLI). 
FLI is a valuable method in various biomedical imaging applications, such as gaining insights into the cellular microenvironment in microbiology and distinguishing tumor margins in cancer imaging \cite{dmitriev2021luminescence,yuan2024antibody,verma2025fluorescence}. However, a key difficulty with FLiDAR in cancer tumor imaging is that the resulting time-resolved signal is a complex convolution of the target's distance and its fluorescence decay characteristics \cite{han2010analytical,miller2017noninvasive,smith2020macroscopic,petusseau2024subsurface}. Therefore, our main goal is to isolate depth and fluorescence lifetime information from time-resolved decay signals (\autoref{fig:overall}).

Traditional approaches attempt to compensate for these scattering-induced distortions, often employing inversion methods that fit signals to predefined basis functions \citep{han2010analytical,miller2017noninvasive,petusseau2024subsurface}. These techniques are typically calibrated to specific optical systems and rely on simplifying assumptions. Consequently, their adaptability across diverse scattering environments is often limited. Given these inherent limitations of analytical approaches, Deep Learning (DL) has emerged as a compelling paradigm. DL offers the potential to learn the non-linear relationships from the raw, distorted temporal signals to the underlying physical parameters like depth and fluorescence lifetime \cite{smith2020macroscopic}. However, such DL models typically process the entire temporal input sequence holistically. This global treatment, often guided by generalized inductive biases that do not align well with the specific physics of LiDAR signals in different media, can lead to an ineffective capture of the signal's rich temporal structure.  To mitigate this, some approaches attempt to incorporate a multi-stage process: first, medium optical properties, such as scattering, are estimated using spatially modulated light, and subsequently, these derived properties serve as refinements for analytical methods \citep{petusseau2024subsurface} or auxiliary inputs to DL models \citep{smith2020macroscopic}. While providing physical context, this initial analytical step often retains limitations such as computational expense and reliance on extra measurements, potentially hindering the end-to-end real-time applicability desired for dynamic FLiDAR applications. Our work, in contrast, develops a fully data-driven, end-to-end trainable system designed for direct parameter estimation from raw temporal signals, offering a more promising pathway towards real-time implementation.

In this context, Mixture-of-Experts (MoE) architectures present a promising paradigm by enabling different expert subnetworks to specialize on distinct aspects of the data. Their practical efficacy, especially for such physically governed signals, heavily relies on how expertise is defined and managed \citep{shazeer2017outrageously,masoudnia2014mixture}. Conventional MoE gating mechanisms, tasked with learning how to route or weight information for different experts, typically operate on the raw input features \citep{masoudnia2014mixture}. Without explicit guidance reflecting the underlying physics, these gating mechanisms may collapse to where a few experts dominate \citep{chi2022representation}. This dominance may fail to isolate the temporal features for estimating depth and lifetime in varying scattering conditions, resulting in an inaccurate estimation of %one, if not both, of 
these physical characteristics. To overcome these limitations and model the underlying signal dynamics, we introduce a Physics-guided MoE model, where expert roles are pre-defined based on the temporal segments of the signal. Specifically, an \emph{`Early Expert'} focuses on initial photon arrivals most indicative of depth \citep{zhao2014p}, a \emph{`Late Expert'} analyzes the subsequent decay characteristics embedding the lifetime \citep{zhao2014p,han2010analytical}, and a \emph{`Global Expert'} integrates information across the entire temporal sequence. 

Despite the advantages of temporally specialized experts, the inherent stochasticity of photons' trajectory in the medium and the variability of scattering media mean that the reliability of any single expert's output can still fluctuate. Precise uncertainty estimation is thus crucial for LiDAR applications. Specifically, this estimation is essential for surgical guidance, where inaccurate depth predictions risk damaging healthy tissues. For instance, an Early Expert might produce less specific depth estimates when initial photon returns are heavily distorted, or a Late Expert's lifetime assessment could be compromised by high background noise or complex, overlapping fluorescence decays. This variability highlights an essential remaining challenge: the need to dynamically and quantitatively assess the quality of each expert's contribution.
% Standard deep learning outputs often lack well-calibrated uncertainty measures, and even specialized experts may not inherently signal their current predictive reliability.  

To address the need for dynamic reliability assessment, we propose an  \emph{EvidenceMoE framework} that leverages a novel evidence-based critic mechanism paired with each specialized expert. These critics act as internal evaluators, assessing the quality of their respective experts' predictions. Each critic also generates a correction signal informed by this uncertainty assessment, allowing for the potential refinement of individual expert outputs. This expert-specific reliability information, generated by the critics, informs the final decision-making process, managed by a component we term the Decider. Its primary role is to produce a final prediction by considering what each expert says and how the critics assess experts' reliability. Hence, allowing us to have reliable expert opinions contributes more significantly to the final, unified estimation of depth and lifetime (\autoref{fig:overall}). 

Our EvidenceMoE framework can be explained as being analogous to a peer-review system. Initially, specialized Experts act as researchers, each submitting a focused `draft' (an initial estimation of depth or lifetime) based on their analysis. Next, each of these `drafts' is evaluated by a paired critic, serving as a reviewer. Each critic assigns a quality score reflecting their confidence in the expert's findings and proposes necessary `revisions' or corrections.
Finally, a Decider network, functioning like an editor, intelligently weighs these critically reviewed and refined contributions based on their assessed credibility and synthesizes them into the final publication, our trustworthy estimate. 

The principal contributions of this work are thus:
\begin{itemize}[leftmargin=*]
    \item  A direct approach to estimate depth and lifetime from raw temporal FLiDAR signals, obviating the need for explicit calculation of medium optical properties and thereby showing strong potential for real-time clinical deployment, especially for cancer diagnostics.
    \item   The introduction of quality scores and correction signals, providing reliability indicators for applications such as surgical guidance, where mitigating risks associated with inaccurate estimations in highly scattering environments is important.
    \item   A novel physics-guided Mixture-of-Experts architecture that enables segment-specific expert processing for FLiDAR signals, and an adaptive fusion module for dynamic expert weighting.
    \item   An Evidence-Based Dirichlet Critic (EDC) mechanism predicting Beta distribution parameters for expert quality assessment and correction signals.
\end{itemize}

\section{Background and Related Work}
Our proposed framework integrates concepts from Mixture-of-Experts, uncertainty quantification, and evidential reasoning, all applied to the specific challenges of interpreting time-resolved FLiDAR signals from scattering media. We briefly review these foundational areas and position our contributions within their context.

\subsection{Physics-Guided Inductive Biases for FLiDAR Temporal Segmentation}

With a finite training set, a model’s ability to generalize to new inputs depends on the preferences or assumptions it encodes, its inductive biases, which narrow the set of solutions consistent with the observed data\cite{goyal2022inductive}. These inductive biases are  the inherent assumptions within a model architecture that guide its learning process and ability to generalize from finite data \citep{mitchell1980need}. For complex data such as FLiDAR signals from scattering media, where distinct temporal segments like early and late photon arrivals convey different physical information \citep{zhao2014p}, the choice of inductive bias substantially impacts learning outcomes. An effective inductive bias guides the learning algorithm by incorporating domain knowledge about the signal, such as by structuring the model to process these physically meaningful segments differentially. This targeted approach constrains the hypothesis space, directing the model to focus on relevant features and relationships that align with known physical principles. Consequently, such guidance can lead to more efficient model training, including faster convergence and improved sample efficiency, as the model is steered away from learning spurious correlations, such as noise, ultimately enhancing its ability to interpret the underlying signal characteristics.

\subsection{Mixture-of-Experts (MoE)}

Mixture-of-Experts (MoE) models \citep{shazeer2017outrageously,masoudnia2014mixture} implement a divide-and-conquer strategy by routing each input to a small subset of expert subnetworks via a gating network. In MoE, the gate selects experts through sparse routing to balance computation across the model. This paradigm has been applied to both multi-representation sensing and temporal forecasting. In LiMoE, features from range images, voxels, and point clouds are fused through an MoE layer for LiDAR perception in air \citep{xu2025limoe}. ME-ODAL applies MoE routing to 3D object detection in point clouds \citep{katkoria2024me}. In time-series forecasting, MoE variants incorporate feature transforms or temporal scaling to handle extended sequences \citep{shi2024time}. Across these applications, expert specialization, where each expert learns a distinct function or handles a particular data subset, underpins model behavior \citep{masoudnia2014mixture}. 

Existing temporal MoE methods process each sequence holistically, relying on auxiliary objectives (e.g., load balancing) to promote expert diversity and expecting the gate to infer segment relevance from raw inputs \citep{mu2025comprehensive}. Such approaches do not exploit known signal physics. Our Physics-guided MoE instead assigns one expert to each of the predefined temporal segments, reflecting the physics of photon transport. This segment-based assignment embeds an inductive bias: each expert models the dynamics specific to its interval. 

\subsection{Limitations of Standard Uncertainty Quantification in Deep Learning}
The challenging nature of FLiDAR-based parameter estimation in scattering media, stemming from distorted and convolved signals, elevates the importance of providing reliable uncertainty estimates alongside deep learning predictions. Common approaches of uncertainty estimation include Monte Carlo Dropout \citep{gal2016dropout}, Deep Ensembles \citep{lakshminarayanan2017simple}, and Bayesian Neural Networks (BNNs) \citep{Blundell2015Weight}. While effective, these methods can suffer from limitations \citep{ovadia2019can}: MC Dropout's estimates may not always be well-calibrated; Deep Ensembles incur significant computational overhead during training and inference \citep{lakshminarayanan2017simple}; and approximate inference in BNNs can be complex to implement and tune \citep{jospin2022hands}. Furthermore, these methods primarily capture aleatoric (data) uncertainty or epistemic (model) uncertainty \citep{kendall2017uncertainties,hullermeier2021aleatoric}, but do not explicitly model the quality of a specific prediction based on input evidence and do not provide correction signals based on the uncertainty.

\subsection{Evidential Deep Learning }

Evidential Deep Learning (EDL) trains a deterministic network, for example, a stack of standard MLP layers, to output the parameters of a higher‐order evidential distribution, rather than placing priors on weights as in BNNs with Bayesian layers \citep{Sensoy2018Evidential,amini2020deep}. Training of EDL models generally relies on loss functions that combine a negative log‐likelihood term with regularizers (e.g.\ KL divergence to a noninformative prior or evidence penalties). Our Evidence-Based Dirichlet Critic (EDC) module acts as an evaluator for each expert. Inspired by actor-critic frameworks in reinforcement learning \citep{grondman2012survey}, though adapted for supervised regression, the critic assesses the actor's (expert's) output. 

\section{Model Architecture}
Our EvidenceMoE framework aims to accurately estimate depth $y_d$ and fluorescence lifetime $y_l$ from time-resolved FLiDAR signals, often compromised by scattering media. Our approach achieves this through a multi-stage system that combines physics-guided expert specialization with an evidence-based mechanism for reliability assessment, output refinement, and adaptive fusion. This section details this architecture, commencing with an overview of its principal components and the overall flow of signal processing. Subsequent subsections will then provide in-depth descriptions of the specialized expert networks, the Evidence-Based Dirichlet Critics (EDCs), and the Decider network.

\subsection{High-Level Overview and Signal Flow}

\begin{figure}
\centering
\includegraphics[width=0.95\textwidth]{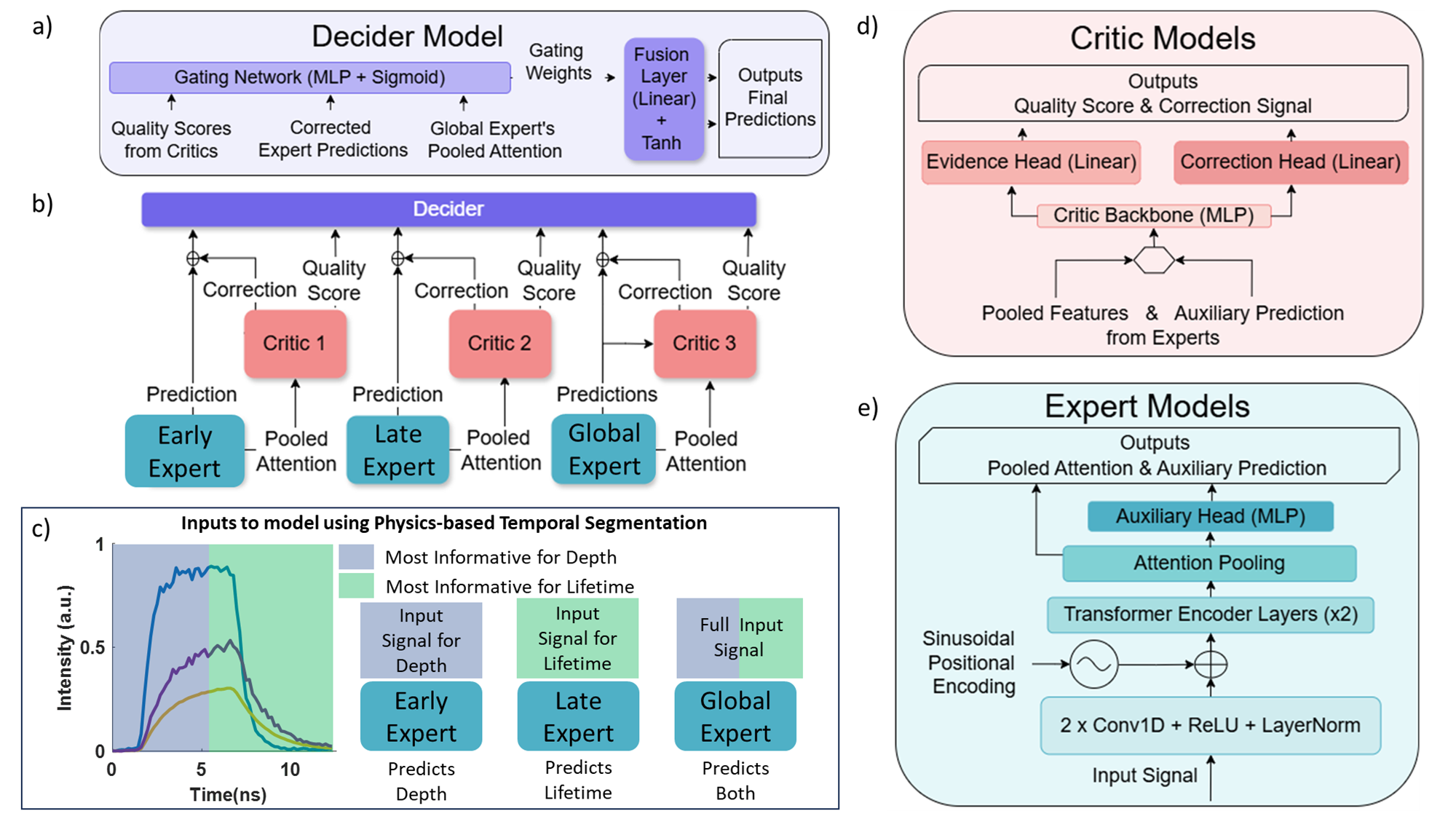}
\caption{Overall architecture of the Physics-guided EvidenceMoE model with Evidence Critics.} %a),b),c),d),e)}
\label{fig:modelarchitecture}
\vspace{-0.2in}
\end{figure}

Our proposed EvidenceMoE model, illustrated in \autoref{fig:modelarchitecture}, processes an input time-resolved FLiDAR signal through a sequence of specialized, interconnected modules. Each input signal is represented as a vector $\vect{x} \in \mathbb{R}^L$, where $L$ denotes the total number of discrete time bins capturing the temporal distribution of detected photons. 

\textbf{Expert pathways} The framework utilizes three parallel expert pathways, each tailored to different aspects of the signal: (1) an Early Expert $E_e$, (2) a Late Expert $E_l$, and (3) a Global Expert $E_g$. These experts process parts ($E_e$ and $E_l$) or the whole ($E_g$) input signal, focusing on features most pertinent to depth or lifetime estimation. 
Each expert $E_k$ generates two outputs: an initial (auxiliary) prediction $y_{\text{aux},k}$ and a pooled feature vector $\phi_k$ derived from an attention pooling stage within the expert.

\textbf{Critic pathways} A key feature of EvidenceMoE is that each expert $E_k$ is paired with a dedicated Evidence-Based Dirichlet Critic (EDC), denoted as $C_k$. The primary function of each critic $C_k$ is to evaluate its corresponding expert's initial prediction $y_{\text{aux},k}$, and pooled feature vector $\phi_k$. 
Based on this assessment, each EDC quantifies the trustworthiness of the expert’s output with a quality score $q_k$ and proposes a corrective signal $\Delta_k$ to refine each expert’s output.

\textbf{Decider network} The expert-specific quality scores, and expert predictions, refined by the corrective signals, ${y}_{\text{aux},k}$, along with the pooled feature vector $\phi_g$ from the Global Expert, are then channeled to the final Decider network $F$. 
%The Decider's role is to synthesize this comprehensive set of inputs. 
The Decider performs an adaptive fusion of the contributions from all experts, where the influence of each refined prediction is dynamically weighted based on its corresponding EDC-derived quality score. On the other hand, the Global Expert's pooled feature vector provides overarching contextual information about the entire signal, aiding the Decider in making more informed fusion decisions.

%\textbf{This, uncertainty-aware feedback from all expert-critic pairs, comprising the potentially corrected expert predictions (e.g., $\tilde{y}_{\text{aux},k}$) and their associated EDC-derived quality scores, is then channeled to a final Decider network, denoted $F$. The Decider $F$ synthesizes these multi-faceted inputs, adaptively fusing the expert contributions based on their evaluated trustworthiness to generate the framework’s unified and refined estimates for depth ($y_d$) and fluorescence lifetime ($y_l$).}

% Conceptually, the signal processing flow within EvidenceMoE can be summarized as follows:

% \begin{enumerate}
%     \item \textbf{Physics-guided mixture-of-experts enable temporal specialization:} The input FLiDAR signal is directed to specialized experts, each focusing on predefined temporal characteristics.
    
%     \item \textbf{Evidence-based Dirichlet critics (EDCs) for reliability assessment and output refinement}: Each expert's output is assessed by its paired EDC, yielding quality scores and correction signals.
    
%     \item \textbf{Decider network for adaptive fusion}\textbf{:} The Decider network synthesizes information from all experts, guided by the EDC-derived reliability metrics, and the global pooled features to produce the final parameter estimations.
% \end{enumerate}

\subsection{Physics-Guided Mixture-of-Experts: Temporal Specialization}
\label{sec:methodology_expert_definitions}
EvidenceMoE's ability to interpret complex FLiDAR signals stems from its specialized MoE layer. This physics-guided component assigns each expert to capture signal features based on known physical principles governing how different types of information are encoded across temporal segments.

\textbf{Expert definitions and temporal assignments}: 
Our proposed MoE architecture leverages the understanding that early-arriving photons in the FLiDAR signal are predominantly correlated with the target depth, whereas the decay characteristics of the later portion of the signal are more significantly influenced by the material's intrinsic fluorescence lifetime. Consequently, each expert network processes a designated portion of the input signal (cf.~\autoref{fig:modelarchitecture}).

\textbf{Implementation details} Each expert $E_k$ shares a common internal architecture designed to process its assigned temporal signal segment $\vect{x}_k$ and produce both an auxiliary prediction $y_{\text{aux},k}$ and a pooled feature vector $\phi_k \in \mathbb{R}^H$. This architecture, detailed in \autoref{sec:expertArchDetails}, employs a hybrid approach: First, it applies 1D convolutional layers to extract local temporal features, followed by a transformer encoder to capture long-range contextual dependencies within the segment. An attention pooling layer then condenses these processed features into the fixed-size vector $\phi_k$, from which a multi-layer perceptron (MLP) head generates the final prediction $y_{\text{aux},k}$.

\subsection{Evidence-Based Dirichlet Critics (EDCs)}
\label{sec:edc}

To account for the stochasticity inherent in photon transport and signal noise in scattering media, EvidenceMoE incorporates Evidence-Based Dirichlet Critics (EDCs) that assess expert reliability and refine outputs. While the physics-guided specialization of experts improves their capacity to model distinct temporal regions of the FLiDAR signal, their predictions can still vary in reliability. The EDCs serve as lightweight evaluators that complement each expert by estimating uncertainty and proposing corrective refinements, enabling a more robust downstream fusion of expert outputs.

\textbf{Critic definition} Each EDC is paired with a specific expert and receives a rich input representation that includes both the expert's pooled internal feature vector and its auxiliary prediction. These are concatenated into a single input vector, allowing the critic to consider both the latent activations and surface-level outputs when evaluating reliability. Formally, the input to critic $C_k$ is given by $z_k = \text{concat}(\phi_k, y_{\text{aux},k})$, 
%\begin{equation}
%    z_k = \text{concat}(\phi_k, y_{\text{aux},k}),
%    \label{eq:critic_input}
%\end{equation}
where $\phi_k \in \mathbb{R}^H$ denotes the expert’s pooled features and $y_{\text{aux},k} \in \mathbb{R}^{D_k}$ the auxiliary prediction, resulting in $z_k \in \mathbb{R}^{H + D_k}$.

The critic architecture consists of a shared MLP backbone that processes the input $z_k$ into a latent representation, which is then routed through two independent heads. The first, an \textit{evidence head}, estimates parameters for a product of independent Beta distributions (one per output dimension) designed to model prediction reliability. The output of this head consists of parameters $\alpha_{k,d}$ and $\beta_{k,d}$ for each prediction dimension $d$, capturing the shape of uncertainty over the expert’s outputs. The second, a \textit{correction head}, produces a residual vector $\Delta_k \in \mathbb{R}^{D_k}$, which can be added to the auxiliary prediction to yield a refined output.

\textbf{Critic score} In the absence of direct supervision, training the evidence head requires a proxy for an expert’s prediction quality. We compute the target quality score as the inverse of the expert’s Mean Absolute Error (MAE) on each output dimension. For each dimension $d$, the target score is given by
\begin{equation}\label{eq:critic_target_quality_method}
    q^{\text{target}}_{k,d} = \frac{1}{1 + \kappa \cdot \text{MAE}_{k,d}} \quad \text{with} \quad \text{MAE}_{k,d} = \frac{1}{N} \sum_{i=1}^{N} \left| y_{{aux},k,d}^{(i)} - y_{{true},k',d}^{(i)} \right|,
\end{equation}

where $N$ represents the batch size and $\kappa$ is a scaling hyperparameter controlling sensitivity. This formulation enables the critic to learn a continuous notion of reliability directly grounded in the expert’s observed performance.

\subsection{Decider Network: Adaptive and Informed Fusion}
\label{sec:decider}

To synthesize the multiple, quality-assessed predictions from the expert pathways into a  unified estimation of depth and lifetime, the EvidenceMoE architecture incorporates a Decider network. The Decider architecture ${F}$, illustrated in \autoref{fig:modelarchitecture} consists of a gating mechanism and a fusion layer.

\textbf{Gating mechanism for dynamic expert weighting}
\label{sec:methodology_gating_mechanism}
A core component of the Decider is a learned gating network $G$.
The gate's decision $w$ is informed by a concatenation of the corrected auxiliary predictions, ${u}_{{gate}} = \text{concat}(y_{{aux},k}, \phi_g, {q}_{full})$, the global decider feature, and the full quality scores ${q}_{full}$, a vector containing the mean quality scores derived from the EDCs for all relevant output dimensions of the experts. The gating network $G$ outputs gating weights, $w_e$, $w_l$, and $w_g$ for early, late, and global experts respectively. 
\begin{equation}
    {w} = \sigma( {W}_{g2} \cdot \text{ReLU}({W}_{g1} {u}_{\text{gate}} + {b}_{g1}) + {b}_{g2} ) \label{eq:decider_gate_weights}
\end{equation}

where $({W}_{g1}, {b}_{g1}, {W}_{g2}, {b}_{g2})$ are learnable parameters within ${F}$. These weights determine the relative influence of each expert branch.

\textbf{Fusion and final prediction}
\label{sec:methodology_decider_fusion_layer}

The gated expert contributions are concatenated with the decider feature $\phi_g$:
\begin{equation}
    {{y}}_{\text{gated}} = [{y}_{\text{aux},e} \cdot w_e, {y}_{\text{aux},l} \cdot w_l, {y}_{\text{aux},g,d} \cdot w_g, {y}_{\text{aux},g,l} \cdot w_g] \in \mathbb{R}^{D_{\text{experts}}}
\end{equation}

This vector of gated expert contributions, ${{y}}_{\text{gated}}$, is then concatenated with the $\phi_{g}$ and passed through a final fusion layer ($H_{{fus}}$), a linear layer, to produce the raw 2D output ${y}_{raw} \in \mathbb{R}^2$:
\begin{equation}
    y_{raw} = H_{{fus}}(\text{concat}(y_{gated}, \phi_g) ; \params{F}) \label{eq:decider_raw_output}
\end{equation}
Finally, this raw output ${y}_{{raw}}$ is transformed via an $\tanh$ activation to yield the final predictions for depth $y_d$ and fluorescence lifetime $y_l$.
% \begin{align}
%     y_d &= {transform}_d(y_{{raw},d}) \label{eq:final_pred_d} \\
%     y_l &= {transform}_l(y_{{raw},l}) \label{eq:final_pred_l}
% \end{align}
% The complete set of parameters for the Decider network, including those for the gating network $G$, fusion layer $H_{\text{fus}}$, and output transformation, are part of $\params{F}$.
\subsubsection{Phased Training Strategy}

\textbf{Phased Training Strategy}
A three-phased training strategy was utilized to stabilize training.
\begin{enumerate}[leftmargin=*]
    \item \textbf{Phase 1 (Expert Pretraining):} Initially, only the expert parameters (${E}$) are trained for a set number of epochs ($N_1$). Here, the primary learning signal comes from the auxiliary loss $\mathcal{L}_{\text{aux}}$ (detailed in \autoref{sec:loss}) associated with each expert, while the critic (${C}$) and decider (${F}$) components remain frozen.
    \item \textbf{Phase 2 (Critic and Decider Integration):} In the second phase, the expert parameters (${E}$) are frozen. The critic (${C}$) and decider (${F}$) parameters are then trained for $N_2$ epochs, utilizing the full composite loss $\mathcal{L}_{{total}}$ (detailed in \autoref{sec:loss}) to learn how to evaluate the pre-trained experts and fuse their outputs.
    \item \textbf{Phase 3 (Joint Training):} Finally, all model parameters (${E}, {C}, {F}$) are jointly fine-tuned for the remaining $N_3$ epochs using the complete loss function $\mathcal{L}_{{total}}$. This phase incorporates a gradual unfreezing schedule for the expert learning rates.
\end{enumerate}
The specific number of epochs allocated to each phase ($N_1, N_2, N_3$) was subject to variation in some experiments, as further detailed in the ablation studies (\autoref{tab:ablation-k2}).
\section{Empirical Studies}

This section details the empirical validation of the EvidenceMoE framework, encompassing the FLiDAR data simulation workflow and evaluation metrics. It highlights the critical role of physics-guided focus on the first and second halves of the temporal FLiDAR signal for accurate depth and lifetime measurement, while effectively managing photon stochasticity using EDCs. We then present the quantitative results of the complete EvidenceMoE framework, followed by detailed ablation studies that assess the contributions of individual architectural components and training strategies. For all experiments, the dataset was partitioned into 80\% training, 20\% validation, and 100 test samples. All experiments were conducted using PyTorch on a high-performance computing cluster, with each node equipped with dual 20-core 2.5 GHz Intel Xeon Gold 6248 CPUs, 768 GiB of RAM, and eight NVIDIA Tesla V100 GPUs (32 GiB HBM each).
% For all experiments, the dataset was partitioned into 80\% for training, 20\% for validation, and 100 test samples.
% All experiments were conducted using PyTorch on a high-performance computing cluster. Each node was equipped with dual 20-core 2.5 GHz Intel Xeon Gold 6248 CPUs, 768 GiB of RAM, and eight NVIDIA Tesla V100 GPUs, each with 32 GiB of HBM. 

\subsection{FliDAR data simulation in scattering media}
To generate realistic FLiDAR data, we employed Monte Carlo (MC) simulation, a robust methodology for simulating photon transport in scattering media like biological tissues \citep{chen2012optical, fang2009monte}, using the Monte Carlo eXtreme (MCX) tool \citep{fang2009monte}. We adopted a two-stage approach, excitation to fluorophore and emission from fluorophore, both within scattering media, to generate time-resolved fluorescence signals, following well-established workflows \citep {nizam2024novel, nizam2024monte, pandey2024deep}. This radiative transport equation-based simulated data inherently captures the physics of light propagation through tissue by utilizing the camera specific characteristics, instrument response function (IRF) and noise profile, which are essential for ensuring accurate stochastic variations and statistical fidelity of photon counts.

\subsection{Evaluation Metrics}

We have employed a suite of metrics targeting different aspects of prediction accuracy to evaluate
the performance of EvidenceMoE and its variants. For depth estimation accuracy, we utilized the
Normalized Root Mean Squared Error (D.NRMSE), the Absolute Relative Error (D.AbsRel), and
the Root Mean Squared Logarithmic Error (D.RMSElog). Fluorescence lifetime estimation accuracy
was primarily assessed using the Normalized Root Mean Squared Error for Lifetime (L.NRMSE(f)).

\subsection{Validating Physics-Guided Expert Specialization}

The Physics-Guided Expert design is rooted in the understanding that distinct temporal segments of the FLiDAR signal carry dominant information for different parameters: early photon arrivals are most indicative of depth, while later decay characteristics primarily encode fluorescence lifetime.

We have analyzed the pooled attention features generated within each expert to validate that physics-guided approach enables effective specialization. These features, derived via an attention pooling mechanism, highlight the temporal regions of the input signal segment that each expert deems most salient for its specific prediction task. 

\begin{figure}[h!]
\centering
\includegraphics[width=0.7\textwidth]{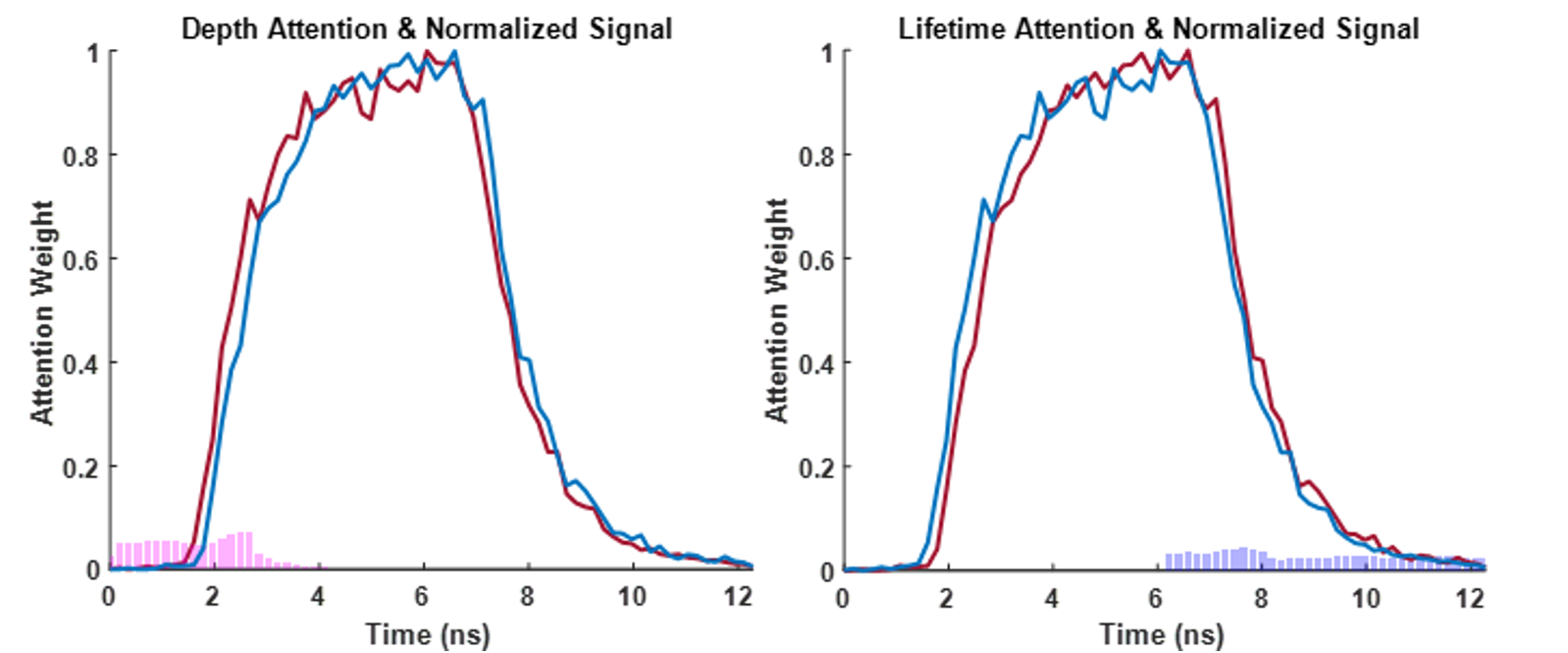}
\caption{Visualization of Pooled Attention Weights Over Time for Both Depth and Lifetime Experts, Computed Over a Batch Size of 512. }
\label{fig:attention}
\vspace{-0.2in}
\end{figure}

Qualitative analysis of attention maps illustrated in \autoref{fig:attention} confirms this intended and effective specialization. The Early Expert consistently assigns higher attention weights to the very early time bins within its designated input segment, corresponding to the rising edge and initial peak of the overall FLiDAR signal. Conversely, the Late Expert's attention is concentrated on the later portions of its respective input segment, capturing the signal's decay tail.

%This observed focused attention validates our physics-guided strategy. By providing each expert with a specific temporal window aligned with known photon physics, the model learns to extract the relevant features for its designated primary target (depth or lifetime). 

\subsection{The Challenge of Stochasticity}

Estimating parameters from FLiDAR signals in scattering media is inherently challenging due to the stochastic nature of photon transport. This randomness introduces significant variability and noise into the detected signals. A common strategy to account for data-dependent uncertainty is to train models to predict not only a point estimate but also an associated variance, often achieved using a heteroscedastic loss function.  

We first conducted experiments where the expert networks were trained in isolation using such a heteroscedastic loss. This configuration, referred to as \emph{Heteroscedastic experts only} in \autoref{tab:ablation-k2} aimed to directly predict both the target parameters and their corresponding uncertainty (variance) for each input signal, without the benefit of the critic or decider mechanisms.

The results underscore the task's difficulty: This approach demanded a significantly longer training period,~ 500 epochs compared to the 70 epochs required for our proposed EvidenceMoE model, to reach convergence, indicating the complexity of learning the underlying signal-parameter relationships. The depth estimation accuracy (D.NRMSE of 0.036) was about 16\% lower compared to our full EvidenceMoE framework. These findings highlight that merely enabling experts to predict their uncertainty is not sufficient to fully address the complexities of FLiDAR signal analysis in scattering media. This motivates the more sophisticated, multi-faceted approach of EvidenceMoE.

\subsection{Performance of the Full EvidenceMoE Framework}

\begin{figure}[h!]
\centering
\includegraphics[width=\textwidth]{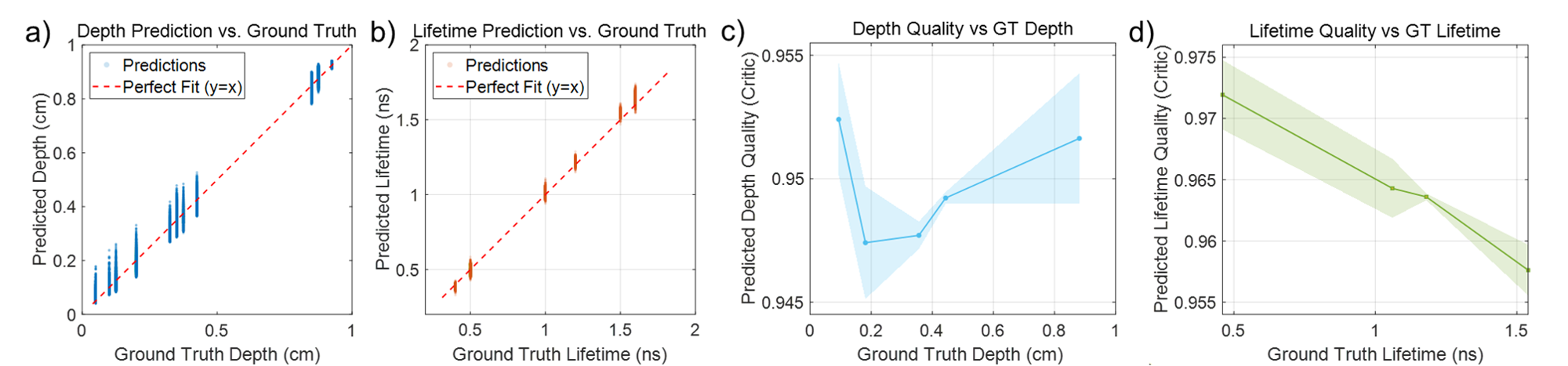}
\caption{Performance results of EvidenceMoE model}
\label{fig:results}
\vspace{-0.2in}
\end{figure}

%Here, we evaluate the performance of the proposed EvidenceMoE framework. 
In our evaluation of our proposed EvidenceMoe model, we set the hyperparameter $\kappa = 2$ (see Equation \ref{eq:critic_target_quality_method}) while training the EDCs. This parameter influences how prediction errors are translated into the target quality scores that the EDCs learn to predict. Our choice of $\kappa = 2 $ was made to ensure a balanced and interpretable relationship between error and quality, where, for example, a 20\% prediction error yields around a 70\% quality score.
As reported in \autoref{tab:ablation-k2} (row \textit{Full model ($\kappa=2$)}), our EvidenceMoE framework demonstrates strong performance in accurately estimating both depth and fluorescence lifetime. The accuracy of these estimations is further illustrated in \autoref{fig:results}, which depicts the predicted depth and lifetime values for 100 test samples against their ground truth values.  Results demonstrate that lifetime predictions (\autoref{fig:results} b) exhibit high precision, closely aligning with the ground truth. While depth predictions (\autoref{fig:results} a) also show strong agreement, they display a slightly larger spread compared to lifetime; yet, the maximum depth errors remain small, around 0.07 cm (0.7 mm), indicating a high degree of accuracy. Notably, the predicted depth quality scores (\autoref{fig:results} c) average around 95\%, while the lifetime quality scores (\autoref{fig:results} d) average around 96.5\%. This observation suggests the slightly higher quality scores for lifetime correspond with its marginally better predictive precision compared to depth, underscoring the utility of the EDC-generated quality scores in reflecting prediction reliability.

\subsubsection{Ablation Studies}
To evaluate the distinct contributions of its architectural elements and design choices, we conducted a series of ablation studies on the EvidenceMoE framework. These studies systematically assessed the impact of individual mechanisms on depth and lifetime estimation by creating model variants with specific components removed or altered. Key aspects investigated included the necessity of evidential correction, quality gating, decider features, and decider fusion. Additionally, we explored the influence of hyperparameter settings, phased training schedules, and pooling mechanisms. Each ablated configuration was evaluated on the same test dataset.
Collectively, these ablation studies (in Table\ref{tab:ablation-k2}) demonstrate the substantial contribution of the integrated physics-guided temporal segmentation, Evidence-Based Dirichlet Critics, and adaptive fusion mechanism to the overall efficacy and robustness of the EvidenceMoE system for FLiDAR-based parameter estimation. 

\begin{table*}
\centering
\scriptsize
\caption{Ablation studies for proposed model.}
\label{tab:ablation-k2}
\begin{tabular}{lcccccc}
\toprule
Configuration                         & D.NRMSE $\downarrow$           & D.AbsRel$\downarrow$             & D.RMSE$_{\log}$ $\downarrow$      & L.NRMSE(f)  $\downarrow$          & Q.Depth   $\uparrow$      & Q.Life   $\uparrow$        \\
\midrule
Damping factor $d=0.5$                & 0.032 $\pm$ 0.009 & 0.164 $\pm$ 0.160 & 0.170 $\pm$ 0.150 & 0.063 $\pm$ 0.023 & 0.954 $\pm$ 0.001 & 0.966 $\pm$ 0.007 \\
Damping factor $d=1.0$                & 0.034 $\pm$ 0.010 & 0.202 $\pm$ 0.240 & 0.189 $\pm$ 0.180 & 0.058 $\pm$ 0.018 & 0.952 $\pm$ 0.004 & 0.966 $\pm$ 0.006 \\
Critic quality‐loss $\lambda_{cq}=4$  & 0.034 $\pm$ 0.011 & 0.150 $\pm$ 0.120 & 0.169 $\pm$ 0.130 & 0.071 $\pm$ 0.022 & 0.949 $\pm$ 0.003 & 0.962 $\pm$ 0.009 \\
Phased training (1 / 6)               & 0.033 $\pm$ 0.009 & 0.171 $\pm$ 0.180 & 0.175 $\pm$ 0.150 & 0.064 $\pm$ 0.014 & 0.951 $\pm$ 0.002 & 0.964 $\pm$ 0.008 \\
Phased training (5 / 15)              & 0.032 $\pm$ 0.008 & 0.153 $\pm$ 0.130 & 0.174 $\pm$ 0.140 & \textbf{0.058 $\pm$ 0.010} & 0.951 $\pm$ 0.003 & 0.962 $\pm$ 0.008 \\
Phased training (3 / 8)               & 0.034 $\pm$ 0.009 & 0.153 $\pm$ 0.130 & 0.166 $\pm$ 0.130 & 0.059 $\pm$ 0.017 & 0.952 $\pm$ 0.002 & 0.968 $\pm$ 0.005 \\
Phased training (10 / 10)             & 0.036 $\pm$ 0.013 & 0.198 $\pm$ 0.200 & 0.383 $\pm$ 0.520 & 0.083 $\pm$ 0.026 & 0.949 $\pm$ 0.000 & 0.963 $\pm$ 0.006 \\
No evidential correction              & 0.035 $\pm$ 0.009 & 0.206 $\pm$ 0.240 & 0.193 $\pm$ 0.180 & 0.063 $\pm$ 0.018 & 0.951 $\pm$ 0.001 & 0.964 $\pm$ 0.007 \\
No quality gating                     & 0.032 $\pm$ 0.009 & 0.143 $\pm$ 0.110 & 0.167 $\pm$ 0.120 & 0.074 $\pm$ 0.028 & 0.950 $\pm$ 0.002 & 0.963 $\pm$ 0.009 \\
No decider features                   & 0.032 $\pm$ 0.006 & 0.160 $\pm$ 0.150 & 0.172 $\pm$ 0.140 & 0.077 $\pm$ 0.030 & 0.947 $\pm$ 0.001 & 0.959 $\pm$ 0.008 \\
No decider fusion                     & 0.034 $\pm$ 0.011 & 0.181 $\pm$ 0.180 & 0.179 $\pm$ 0.150 & 0.059 $\pm$ 0.015 & 0.953 $\pm$ 0.004 & 0.965 $\pm$ 0.006 \\
No gating dropout                     & 0.033 $\pm$ 0.007 & 0.169 $\pm$ 0.170 & 0.171 $\pm$ 0.150 & 0.072 $\pm$ 0.017 & 0.951 $\pm$ 0.001 & 0.961 $\pm$ 0.007 \\
No phased training                    & 0.031 $\pm$ 0.010 & 0.145 $\pm$ 0.120 & 0.159 $\pm$ 0.130 & 0.063 $\pm$ 0.023 & 0.948 $\pm$ 0.002 & 0.964 $\pm$ 0.008 \\
Mean pooling                          & 0.036 $\pm$ 0.010 & 0.170 $\pm$ 0.150 & 0.174 $\pm$ 0.140 & 0.069 $\pm$ 0.024 & 0.949 $\pm$ 0.006 & 0.968 $\pm$ 0.007 \\
Heteroscedastic experts only          & 0.036 $\pm$ 0.011 & 0.139 $\pm$ 0.093 & 0.164 $\pm$ 0.110 & 0.063 $\pm$ 0.051 & -- & -- \\
\midrule
Full model ($\kappa=2$)              & \textbf{0.030 $\pm$ 0.007} & \textbf{0.140 $\pm$ 0.120} & \textbf{0.155 $\pm$ 0.120} & 0.074 $\pm$ 0.022 & 0.950 $\pm$ 0.003 & 0.965 $\pm$ 0.006 \\
\bottomrule
\end{tabular}
\vspace{-0.2in}
\end{table*}

\section{Conclusion}

In this work, we introduced EvidenceMOE, a flexible framework for estimating depth and fluorescence lifetime parameters embedded in time-resolved signals, such as Fluorescence LiDAR. We further proposed a scoring method to quantify prediction reliability. Our expert-critic model architecture reduces instability and achieves robust accuracy across varying signal conditions, with NRMSE over 0.03 and 0.074 for depth and lifetime estimation respectively.
While motivated by fluorescence LiDAR, the approach generalizes to other domains involving complex temporal or spectral signals, such as industrial and biomedical sensing. By delivering both accurate predictions and confidence measures, our framework supports reliable decision-making in high-stakes environments.

\bibliographystyle{plainnat}

\bibliography{references}

\medskip

%%%%%%%%%%%%%%%%%%%%%%%%%%%%%%%%%%%%%%%%%%%%%%%%%%%%%%%%%%%%
\newpage

\section{Principles of Time-Resolved FLiDAR, Photon Scattering, and Fluorescence Lifetime}

Fluorescence LiDAR (FLiDAR) is a specialized technique that extends LiDAR capabilities by analyzing light-induced fluorescence in scattering media, where LiDAR performance degrades by photon scattering. When photons from the laser pulse are absorbed by specific molecules (fluorophores) within the target material, these molecules transition to an excited electronic state. They subsequently relax to their ground state, partly by emitting photons, a process known as \emph{fluorescence} \citep{yuan2024antibody,verma2025fluorescence,dmitriev2021luminescence}. Time-resolved camera systems are designed to detect and temporally resolve this fluorescence emission. The characteristic rate at which the fluorescence intensity declines after excitation follows an exponential curve, and is termed the \emph{fluorescence lifetime} ($\tau$). This lifetime is an intrinsic property of the fluorophore and is sensitive to its local chemical and structural environment \citep{verma2025fluorescence,yuan2024antibody}. Consequently, fluorescence lifetime serves as a useful contrast parameter with applications in biomedical diagnostics to differentiate tissue states (such as healthy or cancerous tissue \citep{mcginty2010wide}) or in environmental science for vegetation analysis \citep{karim2024application}.

Accurately retrieving depth and fluorescence lifetime using FLiDAR in scattering media presents distinct challenges. Firstly, the laser pulse for excitation and basic ranging experiences scattering en route to the target \citep{silva2024exploring,ma2024review}. Secondly, photons that are emitted by the target fluorophores will also propagate through the scattering medium to reach the time-resolved camera. During this transit, these fluorescence photons are subject to similar scattering processes. This additional scattering means that the temporal profile of the fluorescence decay, as recorded by the detector, is not solely governed by the intrinsic fluorescence lifetime of the molecule. Instead, the observed decay profile becomes a convolution of the intrinsic exponential decay with the temporal dispersion effects introduced by photon scattering within the medium.

Therefore, the signal acquired by FLiDAR in such conditions is a composite, reflecting both the target’s range and its fluorescence decay properties, each distorted by scattering. Disentangling these convolved effects to estimate the true depth and the intrinsic fluorescence lifetime accurately requires sophisticated signal processing. Despite these complexities, specific temporal characteristics of the signal provide differential information:
\begin{itemize}
  \item \textbf{Early photons:} Early-arriving photons to the time-resolved camera, having statistically undergone fewer scattering events, correlate more strongly with the shortest path length to the target, thus primarily encoding depth information \citep{zhao2014p}. However, precisely categorizing portions of the signal is challenging because scattering, fluorescence lifetime, and system noise collectively blur the temporal boundary, which may cause partial overlap between depth and fluorescence lifetime information.
  \item \textbf{Late photons:} The decay characteristics of the later portion of the signal are more significantly influenced by the fluorescence lifetime. Yet, this late portion of the signal also remains partially influenced by scattering effects and detector noise, preventing a clear temporal separation solely based on fluorescence decay characteristics. 
\end{itemize}

\section{Detailed Model Architecture for Reproducibility}
\label{sec:expertArchDetails}
\subsection{Physics-Guided Mixture-of-Experts}

Each expert ($E_k$) $k \in \{e, l, g\}$, parameterized by $\theta_E = \{\params{E_e}, \params{E_l}, \params{E_g}\}$,  shares a common internal architecture optimized for feature extraction, consisting of three main stages:
\begin{enumerate}
    \item \textbf{Hybrid CNN-Transformer Encoder:} The input segment ${x}_k \in \R^{L_k}$ is initially processed by a sequence of 1D convolutional layers (kernel size $K$, with residual connections, Layer Normalization, and GELU activation). This allows the network to explicitly capture localized characteristics of the time resolved signal, such as:
        \begin{itemize}
            \item The sharpness and timing of the initial rising edge related to photon arrival time (depth).
            \item Local decay rates or changes in slope within short segments of the fluorescence tail (lifetime).
            \item The presence and shape of small secondary peaks or instrumental response function artifacts within the window.
        \end{itemize}
    This extraction of local waveform motifs precedes the subsequent components. The output of the CNN block is then augmented with sinusoidal Positional Encoding before being passed to a multi-layer Transformer encoder. The Transformer utilizes self-attention mechanisms to model long-range temporal dependencies and contextual relationships across the entire input segment $\vect{x}_k$. The encoder outputs a sequence of refined feature vectors ${h}^{enc}_k \in \R^{L_k \times H}$, where $H$ is the hidden dimension size.
    \item \textbf{Pooling Layer ($f_k$):} To obtain a fixed-size representation from the variable-length output sequence of the encoder, an Attention Pooling mechanism is employed. This layer learns attention weights over the time steps of ${h}^{enc}_k$ and computes a weighted sum, producing a single feature vector $\phi_k \in \R^H$. This vector summarizes the relevant information from the expert's input segment.
    \begin{equation}
        \phi_k = \text{AttentionPool}({h}^{enc}_k; \params{E_k}) \label{eq:expert_pooling}
    \end{equation}
    \item \textbf{Auxiliary Prediction Head ($h_k$):} A small Multi-Layer Perceptron (MLP) with non-linear activation ( ReLU) maps the pooled feature vector $\phi_k$ to the expert's specific auxiliary prediction $y_{aux,k}$ (dimension $D_k=1$ for early/late, $D_k=2$ for global).
    \begin{equation}
        y_{aux,k} = h_k(\phi_k; \params{E_k}) \label{eq:expert_aux_pred_detailed}
    \end{equation}
\end{enumerate}

\textbf{Role of Global Expert Features:}
Beyond its auxiliary prediction $y_{aux,g}$, the pooled feature vector $\phi_g$ from the global expert serves a dual purpose. It acts as the primary source of global context for the downstream fusion mechanism within the Final Decider Head, informing the gating decisions.

\subsection{Evidence-Based Dirichlet Critics (EDC)}
To assess the reliability of each expert's auxiliary prediction and provide a mechanism for refinement, we employ a dedicated critic network $C_k$ for each expert $k \in \{e, l, g\}$, parameterized by $\theta_C = \{\params{C_e}, \params{C_l}, \params{C_g}\}$. We adopt the Evidential Deep Learning (EDL) framework \cite{amini2020deep} to enable the critics to quantify uncertainty in their quality assessments. 

\textbf{Critic Input Features:}
To enable an informed reliability assessment by the critic $C_k$, we provide it with an input representation $z_k$. This input concatenates the pooled feature vector $\phi_k$ from the expert's encoder (\autoref{eq:expert_pooling}) with the expert's auxiliary prediction $y_{aux,k}$ (\autoref{eq:expert_aux_pred_detailed}):

\begin{equation}
    z_k = \text{concat}(\phi_k, y_{aux,k}) \label{eq:critic_input_rich_edc_detailed_again} % Updated label if needed
\end{equation}

\textbf{Critic Architecture:}
Each critic $C_k$ utilizes an identical architecture comprising a shared MLP backbone ($b_k$) followed by two separate linear heads:
\begin{enumerate}
   \item \textbf{Shared Backbone ($b_k$):} The concatenated input $z_k$ is processed by a shared MLP backbone, $b_k$. A relatively shallow architecture (two layers) with moderate hidden dimensions [32, 16] is employed to keep the critic computationally lightweight while providing sufficient capacity to extract relevant features from the input $z_k$. The backbone outputs a shared latent feature representation $h_k \in \R^{16}$.
    \begin{equation}
        h_k = b_k(z_k; \params{C_k})
    \end{equation}
    \item \textbf{Evidence Head ($evi_k$):} A dedicated linear layer, $evi_k$, maps the shared features $h_k$ to the raw evidence outputs $e_k \in \R^{2 \times D_k}$ (one positive and one negative evidence value per output dimension $D_k$ of the corresponding expert).
    \begin{equation}
        e_k = evi_k(h_k; \params{C_k}) \label{eq:raw_evidence_edc_detailed_again}
    \end{equation}
    To ensure the parameters of the resulting Beta/Dirichlet distribution are strictly positive and greater than one (required for a well-defined distribution and stable calculation of variance and KL divergence terms), the raw evidence is transformed using the $\softplus(x) = \log(1 + e^x)$ function followed by adding 1:
    \begin{align}
        \alpha_{k,d} &= \softplus(e_{k, pos, d}) + 1 \label{eq:alpha_edc_detailed_again}\\
        \beta_{k,d} &= \softplus(e_{k, neg, d}) + 1 \label{eq:beta_edc_detailed_again}
    \end{align}
    
    For each dimension $d=1..D_k$, where $e_{k, pos}$ and $e_{k, neg}$ are the corresponding slices of $e_k$. The resulting $\alpha_{k,d}, \beta_{k,d} > 1$ parameterize $D_k$ independent Beta distributions. The total evidence $S_{k,d} = \alpha_{k,d} + \beta_{k,d}$ represents the precision or concentration parameter of the distribution, quantifying the amount of evidence gathered from the data supporting the quality prediction; a higher $S_{k,d}$ indicates greater confidence (lower variance) in the quality estimate.
    \item \textbf{Correction Head ($corr_k$):} A separate linear layer maps $h_k$ to the correction signal $\Delta_k \in \R^{D_k}$.
    \begin{equation}
        \Delta_k = corr_k(h_k; \params{C_k}) \label{eq:correction_edc_detailed_again}
    \end{equation}
\end{enumerate}
The critic's forward pass thus yields $(\alpha_k, \beta_k, \Delta_k)$. The mean of the predicted Beta distribution serves as the point estimate for quality score $q_k$:
\begin{equation}
    q_{k,d} = \frac{\alpha_{k,d}}{\alpha_{k,d} + \beta_{k,d}} \label{eq:mean_quality_edc_detailed_again}
\end{equation}
These mean quality scores (specifically $q_e, q_l,$ and the two components of $q_g$, forming $q_{full} \in \R^4$) are utilized by the downstream decider head.

\subsection{Decider Head and Fusion Mechanism}

The final stage of the model is the Decider Head ($F$, parameterized by $\params{F}$), which performs a learned, context-aware fusion of the information streams originating from the three expert branches. This head utilizes a gating network ($G$) that processes the potentially corrected auxiliary predictions (${y}_{aux,k}$), the critics' reliability estimates ($q_{full}$), and global contextual features ($\phi_{g}$) to compute dynamic, input-dependent weights (${w}$) for each expert branch. These weighted expert contributions are then combined with the global context in a subsequent fusion layer ($H_{fus}$) to generate the final, refined 2D prediction $y_{d}, y_{l}$.

\textbf{Inputs to the Decider Head:}
The head receives three primary inputs per sample:
\begin{enumerate}
    \item \textbf{Corrected Auxiliary Predictions (${y}_{aux}$):} This is the set of auxiliary predictions from the experts,  adjusted by the correction signals provided by the critics (\autoref{eq:correction_edc_detailed_again}), ${y}_{aux} = \{{y}_{aux,e} \in \R^1, {y}_{aux,l} \in \R^1, {y}_{aux,g} \in \R^2\}$.  
    \item \textbf{Decider Feature ($\phi_{g}$):} This is the pooled feature vector $\phi_g \in \R^H$ from the global expert's encoder (\autoref{eq:expert_pooling}).
    \item \textbf{Full Quality Scores ($q_{full}$):} A vector containing the mean quality estimates derived from the Evidence Critics for all four quality dimensions: $q_{full} = [q_e, q_l, q_{g,d}, q_{g,l}] \in \R^4$ (\autoref{eq:mean_quality_edc_detailed_again}). 
\end{enumerate}

\textbf{Gating Network ($G$):}
The core of the fusion mechanism is a learned gating network $G$, designed to adaptively weight the contributions of the three expert branches based on the specific input characteristics.  The gate's decision is informed by the corrected predictions ${y}_{aux}$, the global context $\phi_{g}$, and the critics' full quality assessments $q_{full}$. The combined input $u_{gate}$ is defined as:
\begin{align}
% \tilde{y}_{aux\_concat} &= \text{concat}(\tilde{y}_{aux,e}, \tilde{y}_{aux,l}, \tilde{y}_{aux,g}) \\ % Defined in Inputs
{u}_{{gate}} = \text{concat}(y_{{aux},k}, \phi_g, {q}_{full})
\end{align}
The gating network architecture consists of a two-layer MLP with a ReLU activation after the first layer and a final Sigmoid activation applied after the second layer (mapping to the 3 expert weights):
\begin{equation}
     {w} = \sigma( {W}_{g2} \cdot \text{ReLU}({W}_{g1} u_{gate} +  {b}_{g1}) +  {b}_{g2} ) \label{eq:gate_weights_final_detailed_again}
\end{equation}
where $\sigma(\cdot)$ is the Sigmoid function, and $({W}_{g1},  {b}_{g1}, {W}_{g2},  {b}_{g2})$ are learnable parameters within $\params{F}$. Additionally, gating dropout may apply dropout to $ {w}_g$ during training as a regularization technique.

\textbf{Applying Gates:}
The gating weights $ {w}$ modulate the corrected auxiliary predictions ${y}_{aux,k}$:
\begin{align}
   {y}_{gated,e} &= {y}_{aux,e} \cdot w_e \\
   {y}_{gated,l} &= {y}_{aux,l} \cdot w_l \\
   {y}_{gated,g} &= {y}_{aux,g} \cdot w_g \label{eq:gated_global_final_detailed_again}
\end{align}

\textbf{Fusion Layer ($H_{fus}$):}
The gated expert contributions are concatenated with the decider feature $\phi_g$ and passed through a final linear fusion layer $H_{fus}$ to produce the raw 2D output $y_{raw}$:
\begin{align}
   {y}_{gated\_concat} &= [{y}_{gated,e},{y}_{gated,l},{y}_{gated,g}]  \\
    u_{fus} &= [{y}_{gated\_concat}, \phi_{decider\_fus}] \\
    y_{raw} &= H_{fus}(u_{fus}; \params{F}) \label{eq:raw_pred_final_detailed_again}
\end{align}

\section{Training Methodology}
\label{sec:loss}
\subsection{Loss Function}

The model parameters, partitioned into expert ($\params{E}$), critic ($\params{C}$), and final head ($\params{F}$) groups, are concurrently trained by minimizing a composite, multi-component loss function $\Ls_{total}$. This loss function is designed to achieve several simultaneous objectives: optimizing the final prediction accuracy ($y_{pred}$), encouraging the experts to learn informative intermediate representations ($\phi_k$) and generate reasonable auxiliary predictions ($y_{aux,k}$), training the Evidence Critics to produce quality estimates (represented by $\alpha_k, \beta_k$) and correction signals ($\Delta_k$) that approximate the experts' scaled residual errors ($\Delta_k \approx (y_{true,k'} - y_{aux,k}) / \lambda_{damp}$), and applying specific regularization terms (such as KL divergence for evidence and an evidence-weighted penalty for corrections). The overall loss is a weighted sum of these individual components:
\begin{equation}
    \Ls_{total} = \lambda_{pri} \Ls_{primary} + \lambda_{aux} \Ls_{aux} + \lambda_{crit\_q} \Ls_{quality} + \lambda_{corr} \Ls_{corr} + \lambda_{pen} \Ls_{penalty} 
    \label{eq:total_loss_final_full_revised} % Updated label example
\end{equation}
where the weights $\lambda_{(\cdot)}$ are scalar hyperparameters controlling the relative importance of each term, as defined in the configuration. Each loss component is detailed below.

\textbf{Primary Loss ($\Ls_{primary}$):}
This is the main objective function driving the overall prediction task. It measures the discrepancy between the final fused prediction $y_{pred} \in \R^2$ and the ground truth $y_{true} \in \R^2$ (containing depth and lifetime). We use the Mean Absolute Error (L1 loss):
\begin{equation}
    \Ls_{primary} = \MAE(y_{pred}, y_{true}) = \frac{1}{D_{out}} \sum_{j=1}^{D_{out}} \E_{\text{batch}} \left[ | y_{pred, j} - y_{true, j} | \right]
\end{equation}
where $D_{out}=2$ (depth and lifetime dimensions) and $\E_{\text{batch}}$ denotes the expectation (mean) over the batch samples.

\textbf{Auxiliary Loss ($\Ls_{aux}$):}
To foster expert specialization and enhance training stability, an auxiliary loss term, $\mathcal{L}_{aux}$, provides direct supervision to each individual expert network $E_k$. This loss computes the L1 distance between the expert's auxiliary prediction ($y_{aux,e}$ for depth, $y_{aux,l}$ for lifetime, $y_{aux,g}$ for both) and the corresponding ground truth targets ($y_{true,k'}$). Enforcing this intermediate accuracy encourages each expert to learn representations directly relevant to its specific task domain (temporal segment and target variable). 
\begin{equation}
    \Ls_{aux} = \frac{1}{N_{exp}} \sum_{k \in \{e,l,g\}} \E_{\text{batch}}[\Lone(y_{aux,k}, y_{true,k'})]
\end{equation}
where $N_{exp}=3$.

\textbf{Critic Quality Loss ($\Ls_{quality}$):}
This loss component is responsible for training the evidence head parameters ($\alpha, \beta$) of each Evidence Critic $C_k$. It employs the Evidential Deep Learning (EDL) formulation specifically to learn a calibrated predictive distribution (parameterized by $\alpha_k, \beta_k$) over the quality score associated with the corresponding expert's auxiliary prediction, $y_{aux,k}$. The training objective aims to align the mean of this predicted distribution, $q_k = \alpha_k / (\alpha_k + \beta_k)$, with a target quality score $q_{gt,k}$ derived from the expert's actual error (\autoref{eq:quality_target_invmae_revised}), while simultaneously using the evidential variance and KL divergence terms (Equations \ref{eq:L_evi_revised}, \ref{eq:L_KL_revised}) to ensure the distribution's concentration (total evidence $S_k = \alpha_k + \beta_k$) reflects the true uncertainty or reliability. The target quality $q_{gt,k}$ is calculated using the $y_{aux,k}$:
\begin{equation}\label{eq:critic_target_quality_methodmae}
{MAE}_{k,d} = \frac{1}{N} \sum_{i=1}^{N} \left| y_{{aux},k,d}^{(i)} - y_{{true},k',d}^{(i)} \right|
\end{equation}

\begin{align}
    q_{gt,k,d} &= (1 + \kappa \cdot \text{MAE}_{k,d} + \epsilon)^{-1} \label{eq:quality_target_invmae_revised}
\end{align}
where $\kappa$ is the hyperparameter and $\epsilon$ prevents division by zero.

The quality loss combines the Evidential Regression loss ($\Ls_{evi}$) and a KL divergence regularizer ($\Ls_{KL}$), summed over the four quality dimensions ($k,d$) corresponding to (early, depth), (late, lifetime), (global, depth), (global, lifetime):
\begin{align}
    % Evidential Loss: Matches mean prediction to target and regularizes variance
    \Ls_{evi}(\alpha, \beta, q_{gt}) &= \underbrace{(q_{gt} - \tfrac{\alpha}{\alpha+\beta+\epsilon})^2}_{\text{MSE Term}} + \underbrace{\tfrac{\alpha \beta}{(\alpha+\beta+\epsilon)^2 (\alpha+\beta+1+\epsilon)}}_{\text{Variance Term}} \label{eq:L_evi_revised}\\
    % KL Regularizer: Pulls towards uninformative prior Beta(1,1)
    \Ls_{KL}(\alpha, \beta) &= \KL(\BetaDist(\alpha, \beta) || \BetaDist(1, 1)) \nonumber \\
     &= \lgamma(\alpha+\beta) - \lgamma(\alpha) - \lgamma(\beta) \nonumber \\
     &\quad + (\alpha - 1)(\digamma(\alpha) - \digamma(\alpha+\beta)) \nonumber \\
     &\quad + (\beta - 1)(\digamma(\beta) - \digamma(\alpha+\beta)) \label{eq:L_KL_revised} \\
    % Combined Quality Loss: Averaged over batch and summed over dimensions
    \Ls_{quality} &= \E_{\text{batch}} \left[ \sum_{k,d} \left( \Ls_{evi}(\alpha_{k,d}^{loss}, \beta_{k,d}^{loss}, q_{gt,k,d}) + \lambda_{KL} \max(0, \Ls_{KL}(\alpha_{k,d}^{loss}, \beta_{k,d}^{loss})) \right) \right] \label{eq:loss_quality_final_revised}
\end{align}
The $\Ls_{evi}$ term drives the mean predicted quality towards the target while penalizing high confidence for incorrect predictions via the variance term. The $\Ls_{KL}$ term regularizes the learned distribution, preventing collapse and encouraging uncertainty quantification. Gradients from $\Ls_{quality}$ only update critic parameters $\params{C}$.

\textbf{Correction Loss ($\Ls_{corr}$):}
This loss component trains the correction head of each critic $C_k$ to output a signal $\Delta_k$ (\autoref{eq:correction_edc_detailed_again}) that aims to reduce the error in the expert's auxiliary prediction $y_{aux,k}$. Specifically, it minimizes the Huber loss between the damped corrected prediction ($y_{aux,k} + \lambda_{damp} \Delta_k$) and the ground truth target $y_{true,k'}$.
\begin{equation}
    \Ls_{corr} = \E_{\text{batch}} \left[ \sum_{k \in \{e, l, g\}} \Huber(y_{aux,k} + \lambda_{damp} \Delta_k, y_{true,k'}) \right] \label{eq:loss_corr_final_revised_detailed}
\end{equation}
This loss is calculated using the auxiliary predictions $y_{aux,k}$ and correction signals $\Delta_k$ generated during the standard forward pass. Consequently, the gradients derived from $\Ls_{corr}$ backpropagate to update both the critic parameters $\params{C_k}$ (through $\Delta_k$) and the expert parameters $\params{E_k}$. 
\begin{equation}
    \Ls_{corr} = \frac{1}{N_{exp}} \sum_{k \in \{e, l, g\}} \E_{\text{batch}} \left[ \Huber(y_{aux,k} + \lambda_{damp} \Delta_k, y_{true,k'}) \right] \label{eq:loss_corr_final_revised}
\end{equation}

\textbf{Evidence Penalty Loss ($\Ls_{penalty}$):}
To further regulate the correction mechanism and promote robustness, we use an evidence-weighted penalty term, $\Ls_{penalty}$. This loss component links the confidence of the critic's quality assessment (as measured by the total evidence $S_{k,d}^{loss} = \alpha_{k,d}^{loss} + \beta_{k,d}^{loss}$) to the magnitude of the correction signal $\Delta_k^{loss}$ it proposes. The rationale is to discourage the critic from making large adjustments to the expert's prediction when its own assessment of the expert's quality is uncertain (i.e., when the evidence $S$ is low). This is achieved by penalizing the squared L2 norm of the correction signal, inversely weighted by the total evidence:
\begin{equation}
    \Ls_{penalty} = \E_{\text{batch}} \left[ \sum_{k,d} \gamma \frac{(\Delta_{k,d}^{loss})^2}{S_{k,d}^{loss} + \epsilon} \right] \label{eq:loss_penalty_final_detailed}
\end{equation}
where $\gamma$ is the hyperparameter, the sum is over the relevant output dimensions $(k,d)$, and $\epsilon$ ensures numerical stability. This loss is calculated using the parameters $\alpha^{loss}, \beta^{loss}, \Delta^{loss}$ derived from the critic inputs ($z_k^{loss}$), ensuring that its gradients only update the critic parameters $\params{C}$.  This evidence-aware regularization encourages more cautious corrections under uncertainty compared to standard L2 regularization on $\Delta_k$ alone.

\newpage
\section{Additional Ablation Study ($\kappa = 8$)}
\begin{table}[ht!]
\centering
\caption{Ablation study for $\kappa=8$.}
\label{tab:ablation-k8}
\resizebox{\textwidth}{!}{%
\begin{tabular}{lcccccc}
\toprule
Configuration                         & D.NRMSE $\downarrow$           & D.AbsRel$\downarrow$             & D.RMSE$_{\log}$ $\downarrow$      & L.NRMSE(f)  $\downarrow$          & Q.Depth   $\uparrow$      & Q.Life   $\uparrow$        \\
\midrule
Critic quality‐loss $\lambda_{cq}=4$  & 0.033 $\pm$ 0.011 & 0.146 $\pm$ 0.110 & 0.181 $\pm$ 0.130 & 0.058 $\pm$ 0.012 & 0.856 $\pm$ 0.025 & 0.900 $\pm$ 0.034 \\
No evidential correction              & 0.036 $\pm$ 0.015 & 0.148 $\pm$ 0.100 & 0.182 $\pm$ 0.130 & 0.060 $\pm$ 0.011 & 0.859 $\pm$ 0.012 & 0.869 $\pm$ 0.044 \\
No quality gating                     & 0.033 $\pm$ 0.010 & 0.146 $\pm$ 0.110 & 0.176 $\pm$ 0.130 & 0.056 $\pm$ 0.012 & 0.868 $\pm$ 0.024 & 0.886 $\pm$ 0.031 \\
No decider features                   & 0.032 $\pm$ 0.011 & 0.146 $\pm$ 0.110 & 0.184 $\pm$ 0.130 & 0.067 $\pm$ 0.025 & 0.865 $\pm$ 0.014 & 0.892 $\pm$ 0.037 \\
No decider fusion                     & 0.035 $\pm$ 0.013 & 0.167 $\pm$ 0.160 & 0.182 $\pm$ 0.140 & 0.073 $\pm$ 0.015 & 0.828 $\pm$ 0.020 & 0.879 $\pm$ 0.022 \\
Uniform gating                        & 0.034 $\pm$ 0.010 & 0.157 $\pm$ 0.120 & 0.191 $\pm$ 0.150 & 0.066 $\pm$ 0.018 & 0.861 $\pm$ 0.021 & 0.879 $\pm$ 0.034 \\
No gating dropout                     & 0.041 $\pm$ 0.014 & 0.215 $\pm$ 0.180 & 0.373 $\pm$ 0.400 & 0.073 $\pm$ 0.022 & 0.865 $\pm$ 0.015 & 0.866 $\pm$ 0.029 \\
No phased training                    & 0.037 $\pm$ 0.015 & 0.181 $\pm$ 0.150 & 0.269 $\pm$ 0.260 & 0.076 $\pm$ 0.021 & 0.862 $\pm$ 0.007 & 0.890 $\pm$ 0.037 \\
Mean pooling                          & 0.034 $\pm$ 0.009 & 0.155 $\pm$ 0.130 & 0.173 $\pm$ 0.130 & 0.075 $\pm$ 0.027 & 0.818 $\pm$ 0.069 & 0.866 $\pm$ 0.058 \\
No auxiliary MAE                      & 0.037 $\pm$ 0.014 & 0.172 $\pm$ 0.130 & 0.234 $\pm$ 0.180 & 0.063 $\pm$ 0.018 & 0.850 $\pm$ 0.009 & 0.850 $\pm$ 0.034 \\
\midrule
Model trained with ($\kappa=8$)       & 0.040 $\pm$ 0.013 & 0.216 $\pm$ 0.230 & 0.202 $\pm$ 0.180 & \textbf{0.055 $\pm$ 0.011} & 0.857 $\pm$ 0.023 & 0.896 $\pm$ 0.023 \\
Full model ($\kappa=2$)              & \textbf{0.030 $\pm$ 0.007} & \textbf{0.140 $\pm$ 0.120} & \textbf{0.155 $\pm$ 0.120} & 0.074 $\pm$ 0.022 & 0.950 $\pm$ 0.003 & 0.965 $\pm$ 0.006 \\
\bottomrule
\end{tabular}
}
\end{table}

%%%%%%%%%%%%%%%%%%%%%%%%%%%%%%%%%%%%%%%%%%%%%%%%%%%%%%%%%%%%

\newpage

\end{document}